\documentclass[sigconf]{acmart}

\usepackage{graphicx}
\usepackage{amsmath}
\usepackage{amssymb}
\usepackage{amsthm}
\usepackage{booktabs}
\usepackage{algorithm}
\usepackage{algorithmic}
\usepackage{enumerate}
\usepackage{bbm}
\usepackage{multirow}
\usepackage[FIGTOPCAP]{subfigure}
\def\BibTeX{{\rm B\kern-.05em{\sc i\kern-.025em b}\kern-.08emT\kern-.1667em\lower.7ex\hbox{E}\kern-.125emX}}
    
 \setcopyright{acmcopyright}

\begin{document}
\fancyhead{}
%
\title{Cycle-consistent Conditional Adversarial Transfer Networks}

\author{Jingjing Li}
\affiliation{\institution{University of Electronic Science and Technology of China}\institution{The University of Queensland}}
\email{lijin117@yeah.net}

\author{Erpeng Chen}
\affiliation{\institution{University of Electronic Science and Technology of China}}
\email{cep1126@163.com}

\author{Zhengming Ding}
\affiliation{\institution{Indiana University-Purdue University Indianapolis}}
\email{zd2@iu.edu}

\author{Lei Zhu}
\affiliation{\institution{Shandong Normal University}}
\email{leizhu0608@gmail.com}

\author{Ke Lu}
\affiliation{\institution{University of Electronic Science and Technology of China}}
\email{kel@uestc.edu.cn}

\author{Zi Huang}
\affiliation{\institution{The University of Queensland}}
\email{huang@itee.uq.edu.au}
\renewcommand{\shortauthors}{Jingjing Li et al.}
\renewcommand{\shorttitle}{Cycle-consistent Conditional Adversarial Transfer Networks}

%
\begin{abstract}
Domain adaptation investigates the problem of cross-domain knowledge transfer where the labeled source domain and unlabeled target domain have distinctive data distributions. Recently, adversarial training have been successfully applied to domain adaptation and achieved state-of-the-art performance. However, there is still a fatal weakness existing in current adversarial models which is raised from the equilibrium challenge of adversarial training. Specifically, although most of existing methods are able to confuse the domain discriminator, they cannot guarantee that the source domain and target domain are sufficiently similar. In this paper, we propose a novel approach named {\it cycle-consistent conditional adversarial transfer networks} (3CATN) to handle this issue. Our approach takes care of the domain alignment by leveraging adversarial training. Specifically, we condition the adversarial networks with the cross-covariance of learned features and classifier predictions to capture the multimodal structures of data distributions. However, since the classifier predictions are not certainty information, a strong condition with the predictions is risky when the predictions are not accurate. We, therefore, further propose that the truly domain-invariant features should be able to be translated from one domain to the other. To this end, we introduce two feature translation losses and one cycle-consistent loss into the conditional adversarial domain adaptation networks. Extensive experiments on both classical and large-scale datasets verify that our model is able to outperform previous state-of-the-arts with significant improvements.
\end{abstract}

%
%


\copyrightyear{2019} 
\acmYear{2019} 
\acmConference[MM '19]{Proceedings of the 27th ACM International Conference on Multimedia}{October 21--25, 2019}{Nice, France}
\acmBooktitle{Proceedings of the 27th ACM International Conference on Multimedia (MM '19), October 21--25, 2019, Nice, France}
\acmPrice{15.00}
\acmDOI{10.1145/3343031.3350902}
\acmISBN{978-1-4503-6889-6/19/10}
%

%

%
\maketitle

\section{Introduction}
The amount of data we produce every day is truly mind-boggling, reported by Forbes, there are 2.5 quintillion bytes of data created each day at our current pace. These data consists of not only what we are familiar with but also novel and unseen instances. Conventional machine learning paradigms commonly assume that the test data are seen during the training stage~\cite{xian2017zero,li2019zero,Li_2019_CVPR}. Thus, it is challenging to handle the novel and unseen data in reality with conventional methods. As a practical alternative, transfer leaning~\cite{pan2010survey,li2018transfer,li2019locality} has been verified to be crucial for the success in novel environment. One of the most essential topics of transfer learning is domain adaptation~\cite{pan2011domain,tzeng2014deep,long2018conditional,tzeng2017adversarial,ding2018semi} which investigates the problem of cross-domain knowledge transfer where the labeled source domain and unlabeled target domain have distinctive data distributions.

\begin{figure}[t]
\begin{center}
\includegraphics[width=0.81\linewidth]{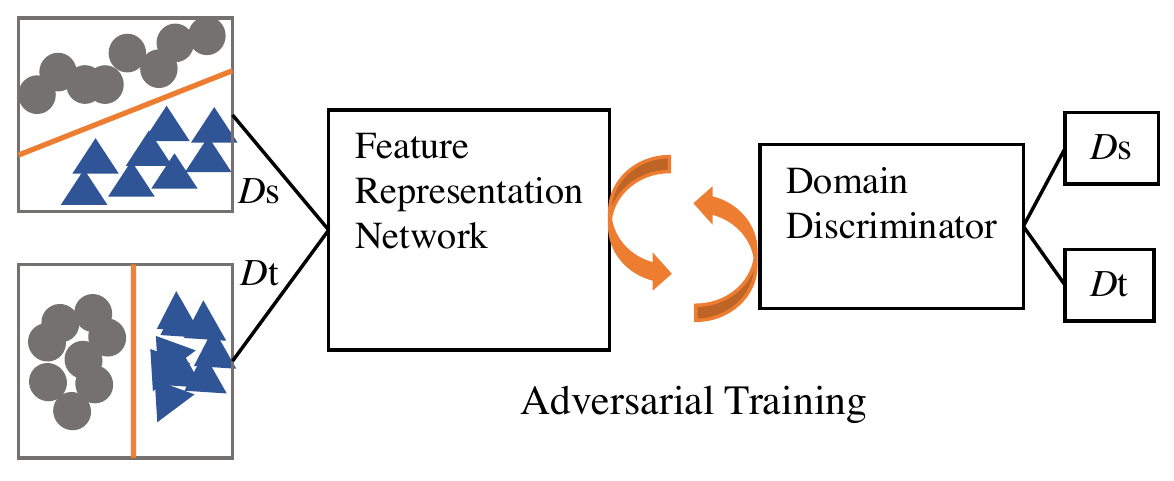}
\end{center}
\vspace{-10pt}
\caption{\small Adversarial domain adaptation networks. 1) The source domain $D_s$ and the target domain $D_t$ have different data distributions. 2) A domain discriminator is trained to distinguish source domain features from target domain features. 3) The feature representation network tries to fool the domain discriminator.}
\label{fig:intro}
\vspace{-20pt} 
\end{figure} 

The very objective of domain adaptation is to reduce the domain distribution shifts between the source domain data and target domain data, so that the models trained on the well labeled source domain can be adapted to the target domain. Existing domain adaptation methods can be categorized into either shallow branch~\cite{pan2011domain,gong2012geodesic,li2018transfer,saenko2010adapting,li2019locality} or deep branch~\cite{ganin2016domain,long2018conditional,hoffman2018cycada,bousmalis2017unsupervised}. Typically, methods in the shallow branch are built as two-step formulations. The first step is used for feature extraction and the second step focuses on domain alignment. Deep methods, however, incorporate the feature extraction and domain alignment into a united architecture. Since the feature extraction component can receive feedbacks from the domain alignment component and reinforce itself, deep approaches achieved state-of-the-art results in recent years~\cite{long2018conditional,hoffman2018cycada}.

Among various deep domain adaptation methods, adversarial domain adaptation~\cite{ganin2016domain,tzeng2017adversarial,hong2018conditional,long2018conditional} attracted a lot of attention lately. Adversarial domain adaptation, as shown in Fig.~\ref{fig:intro}, introduces adversarial training into the deep architecture with the similar idea of generative adversarial networks (GANs)~\cite{goodfellow2014generative,mirza2014conditional}. Specifically, adversarial domain adaptation trains a domain discriminator which aims to distinguish whether a feature is from the source domain or the target domain. At the same time, a deep feature representation model tries to learn domain-invariant features to fool the domain discriminator. Once the domain discriminator fails to tell the feature sources, adversarial domain adaptation assumes that the learned features are well aligned. Then, a classifier trained on the source domain features can be directly deployed to the target domain features~\cite{tzeng2017adversarial,sankaranarayanan2018generate}.

The basic idea of adversarial domain adaptation is sound and it has also been empirically verified to be effective in many applications~\cite{chen2017no,bousmalis2017unsupervised,hoffman2018cycada}. However, there is still a fatal weakness existing in current adversarial models which is raised from the equilibrium challenge of adversarial training~\cite{arora2017generalization}. Specifically, although most of existing methods are able to confuse the domain discriminator, they cannot guarantee that the source domain and target domain are sufficiently similar~\cite{arora2017generalization}. As recently claimed by Long et al.~\cite{long2018conditional}, adversarial adaptation methods may fail to capture the complex multimodal structures which is reflected only by the cross-covariance between the features and corresponding classifier predictions~\cite{song2009hilbert,long2018conditional}. 

In this paper, we propose a novel approach named {\it cycle-consistent conditional adversarial transfer networks} (3CATN) to align the two domains. At first, inspired by very recent work~\cite{long2018conditional}, we deploy a conditional domain discriminator by leveraging the cross-covariance of learned features and corresponding classifier predictions, so that the complex multimodal structures embedded in the data can be captured. However, since the classifier predictions are not certainty information, {\it deploying a strong condition from the classifier is very risky when the predictions are not sufficiently accurate}. If we put an inaccurate condition on the adversarial domain adaptation networks, the learned features may be able to confuse the domain discriminator, but there is no guarantee that the features are truly domain-invariant. Therefore, we further argue that {\it truly domain-invariant features should be able to be translated from one domain to the other}. This assumption does make sense by considering that {\it domain-invariant} actually indicates the feature space is {\it shared} by the two domains. In other words, the domain-invariant features are represented by the same bases and thus can be represented by each other. To this end, we  train two feature translators, one translates features from the source domain to the target domain and the other translates features from the target domain to the source domain. In addition, we calculate a cycle-consistent loss by leveraging the two feature translators. With such a formulation, our method is able to not only capture the complex multimodal data structures but also avoid the negative effects caused by inaccurate conditions. In summary, we list the main contributions of this paper as follows:


\begin{enumerate}[1)]
\item We propose a novel deep method named {\it cycle-consistent conditional adversarial transfer networks} (3CATN) for domain adaptation by taking advantage of adversarial training. Compared with existing adversarial domain adaptation methods, our approach is able to not only capture the complex multimodal structures but also avoid the negative effects caused by uncertain conditions.
\item We argue that the condition with classifier predictions is risky when the the predictions are not sufficiently accurate. To address this, we further propose that truly domain-invariant features should be able to be translated from one domain to the other.
\item Experiments on both classical and large-scale datasets verify that our method is able to outperform previous state-of-the-arts with significant advantages. We show that our method is especially remarkable on relatively hard-to-transfer tasks (predictions on target data are less accurate), which verifies the claims in 2).
\end{enumerate}

The rest of this paper is organized as follows. In section~2, we give a brief review of previous work reported in the community. In section~3, we present the proposed method. Section~4 
reports the experiments. At last, we draw the conclusion in section~5.

\section{Related Work}

\subsection{Transfer Learning and Domain Adaptation}
Domain adaptation~\cite{pan2011domain,ganin2014unsupervised,li2016joint,bousmalis2017unsupervised,long2018conditional,li2019locality} is a popular branch of transfer learning~\cite{pan2010survey,li2018transfer} which was proposed to handle the training data scarcity issue in supervised learning. Since most successful machine learning models~\cite{he2016deep,krizhevsky2012imagenet} are trained in a supervised fashion, it is challenging to handle new environment where there are no sufficient labeled data~\cite{xian2017zero,li2017two}. At the same time, it is also unwise to build every model from scratch with plenty of related models are available~\cite{pan2010survey}. Transfer learning leverages available resources to challenge new situations. In fact, the very widespread success of fine-tuning on pre-trained deep models~\cite{he2016deep,krizhevsky2012imagenet} has verified the effectiveness of transfer leaning.  

A common assumption behind conventional machine learning algorithms is that the training set and test set have identical data distributions. However, this assumption does not always hold in real-world applications~\cite{pan2010survey,visda2017}. Domain adaptation~\cite{pinheiro2018unsupervised,taigman2016unsupervised} was proposed to reduce the domain shifts between the labeled source domain and unlabeled target domain. Early domain adaptation methods~\cite{pan2011domain,gong2012geodesic,aljundi2015landmarks} focus on knowledge transfer techniques which are invariant with samples features. For instance, most of early domain adaptation approaches~\cite{pan2011domain,gong2012geodesic,ding2014latent} learn a common subspace which is shared by the two domains, some of them~\cite{aljundi2015landmarks,hubert2016learning} reweight samples according to the distribution distance with target domain.

With the sweeping success of deep learning, neural networks are also deployed in domain adaptation. At first, deep domain adaptation approaches align the two domains by minimizing a pre-defined metric which measures the feature distributions, e.g., minimizing MMD distance~\cite{long2017deep}, minimizing covariance distance~\cite{sun2016deep}. Recently, adversarial training~\cite{tzeng2017adversarial,ganin2016domain,long2018conditional} was introduced into domain adaptation. The main idea of adversarial domain adaptation is to train a domain discriminator which can distinguish whether a feature is from the source domain or the target domain. The learned features would be considered as domain-invariant once the domain discriminator is confused by the learned features. 

\begin{figure*}[t]
\begin{center}
\includegraphics[width=0.86\linewidth]{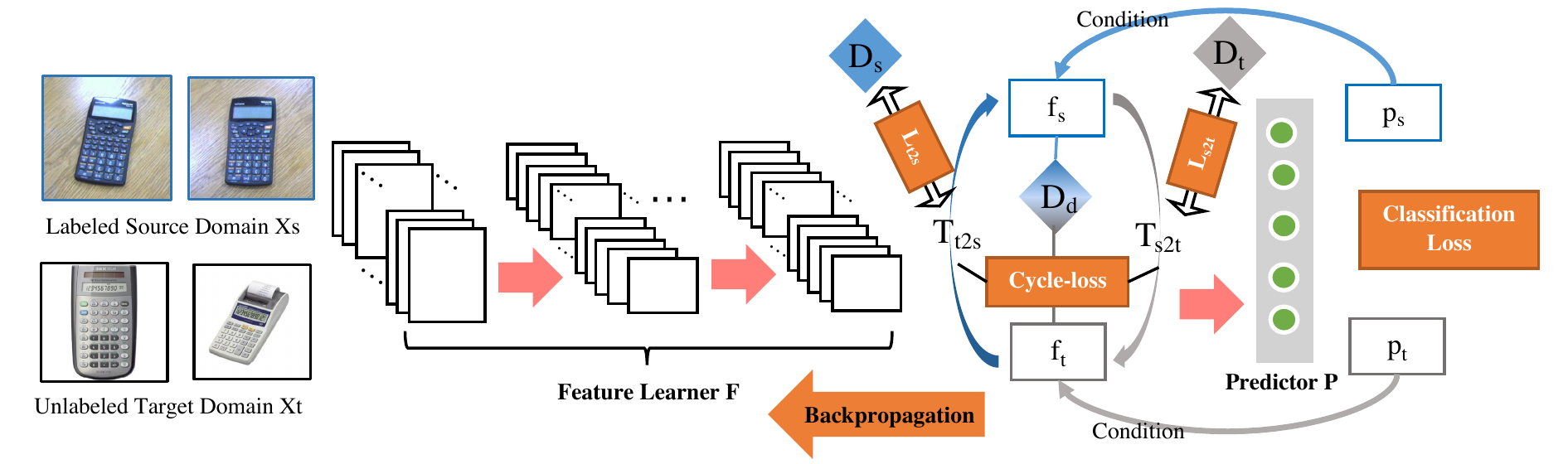}
\end{center}
\vspace{-10pt}
\caption{Idea illustration of our 3CATN. 1) A deep CNN, e.g., ResNet, is trained as a feature learner to learn domain-invariant feature representations. 2) A domain discriminator $D_d$ is trained to distinguish source domain features from target domain features. 3) Two feature translator $T_{s2t}$ and $T_{t2s}$ along with their corresponding discriminators $D_t$ and $D_s$ are trained to translate features from one domain to the other. 4) By leveraging the two feature translators, we calculate a cycle-loss to preserve the translation consistency. 5) The classifier predictions $\mathrm{p}_s$ and $\mathrm{p}_t$ are deployed to condition the adversarial domain adaptation networks. 6) All losses are backpropagated to the feature representation network to learn truly domain-invariant features.}
\label{fig:ideaill}
\end{figure*}

\subsection{Adversarial Domain Adaptation}  
Adversarial domain adaptation~\cite{pan2011domain,gong2012geodesic,ding2014latent,long2018conditional} is similar with generative adversarial networks (GANs)~\cite{goodfellow2014generative} but not exactly the same. GANs typically have a generator and a discriminator where the generator synthesizes fake samples from noises to fool the discriminator. Adversarial domain adaptation, however, usually does not have a ``generator'' which can synthesize something new. The role of ``generator'' in adversarial domain adaptation is played by a feature extractor. Specifically, the feature extractor learns new feature representations which is able to confuse the domain discriminator. Instead of generating from random noises in GANs, the feature extractor learns meaningful feature representations from both source domain data and target domain data.

The adversarial training can be performed at either feature level or pixel level. Feature adaptation methods~\cite{ganin2016domain,tzeng2017adversarial,long2018conditional} minimize the domain discrepancy on the feature space. For instance, Ganin et al.~\cite{ganin2016domain} optimize the standard minimax objective on the feature space. Tzeng et al.~\cite{tzeng2017adversarial} optimize an inverted label objective in a unified framework. Pixel adaptation methods~\cite{bousmalis2017unsupervised,liu2016coupled} propose that the learned representations can be recovered back to raw images. As a result, pixel adaptation methods are much easier to be understood by eyes and more friendly to the end visual tasks. For instance, CoGANs~\cite{liu2016coupled} learns a joint distribution of multi-domain images. However, pixel adaptation methods generally cost much much more than feature-based ones. They are also sensitive to noises and missing pixels. For most classification or segmentation tasks, especially for large-scale applications, it is also unnecessary to directly handle raw pixels. Meaningful features have more discrimination power than pixels. It is worth noting that one of major differences between our method and Hoffman et al.~\cite{hoffman2018cycada} is that the latter emphasizes on pixel adaptation while our method leverages feature translation. Thus, our method is more computational efficient than~\cite{hoffman2018cycada}. In addition, as claimed by Long et al.~\cite{long2018conditional}, the pixel-level adaptation methods are carefully tailored to the digits and synthetic to real adaptation tasks.


In adversarial domain adaptation networks, the idea of conditional GANs~\cite{mirza2014conditional} is widely used to promote the discriminative ability of both feature extractor and domain discriminator. For instance, Chen et al.~\cite{chen2017no} propose a global and class-specific domain adversarial learning framework for road scene segmenters. Long et al.~\cite{long2018conditional} propose two novel strategies to condition the domain discriminator with classifier predictions. Our work is related with CDAN~\cite{long2018conditional}. We also condition our features with the classifier predictions to preserve the multimodal data structures. However, our method is also significantly different from CDAN. The major difference is that CDAN did not consider the situation where the classifier predictions may be not sufficiently accurate (Although entropy condition is proposed in CDAN, entropy itself is an inaccurate metric to measure the classification result). As a result, putting a wrong condition on the discriminator will lead the feature extractor into risky situations. This situation is common in fine-grained visual classification tasks. For instance, a pickup truck can be mis-predicted as a car. With the condition of car on the discriminator, pickups would be classified as cars for good. What is worse, the errors caused by the predictions can be propagated and magnified. In this paper, we argue that truly domain-invariant features should be able to be translated from one domain to the other. In our model, therefore, we introduce two feature translators to map features from one domain to the other. In addition, a cycle-consistent loss is calculated to guarantee that the source samples translated to target style can be translated back to the source domain. Experiments verify that such a formulation is effective for domain adaptation, especially for relatively difficult tasks where the gap between the source domain and the target domain is significantly large. In other words, compared with CDAN, our formulation is able to not only capture the complex multimodal data structures but also prevent and penalize the errors caused by classifier predictions.


\section{The Proposed Method}
\subsection{Notations and Definitions}
In this paper, we use subscripts $s$ and $t$ to denote source and target, respectively. Thus, we have $n_s$ labeled source samples $\{X_s,Y_s\}$ and $n_t$ unlabeled target samples $X_t$ for unsupervised domain adaptation. The main challenge of domain adaptation is that the data distributions, both marginal distributions $\mathcal{P}_s(X_s), \mathcal{P}_t(X_t)$ and conditional distributions $\mathcal{P}_s(Y_s|X_s), \mathcal{P}_t(Y_t|X_t)$ are distinctive from the source domain to the target domain. Therefore, our goal is to train a deep network $F:x\rightarrow y$ which is able to learn discriminative domain-invariant features where the domain shifts can be sufficiently reduced. We define the related concepts as follows:


{\bf Definition 1:} ({Domain}) A domain $\mathcal{D}$ is defined as a sample set $X$ along with its probability distribution $\mathcal{P}(X)$, i.e, $\mathcal{D}=\{X,\mathcal{P}(X)\}$. The source domain and target domain are denoted as $\mathcal{D}_s$ and $\mathcal{D}_t$.

{\bf Definition 2:} (Adversarial Discriminator) An adversarial discriminator $D$ is defined as a binary classifier which is able to distinguish real from fake. In this paper, we introduce a domain discriminator $D_d$ which distinguishes a source domain feature from the target domain. We also train a target sample discriminator $D_t$ to distinguish the real target samples and fake target samples translated from the source instances, and a source sample discriminator $D_s$ to distinguish the real source samples and fake source samples translated from the target domains.

{\bf Definition 3:} (Feature Learner) A feature learner $F$ is defined as a deep neural network which is able to learn domain-invariant feature representations for both the the source and target domains. We denote the feature representation of sample $x$ as $\mathrm{f}=F(x)$.

{\bf Definition 4:} (Feature Predictor) A feature predictor $P$ is defined as a classifier layer which is trained to predict possible categories of a feature representation, i.e, $\mathrm{p}=P(\mathrm{f})$.

{\bf Definition 5:} (Feature Translator) A feature translator $T$ is defined as a mapping which is able to translate the features from one domain to the style of the other domain. In this paper, we train two feature translators $T_{s2t}$ and $T_{t2s}$. 


\subsection{Overall Idea}
Our cycle-consistent conditional adversarial network consists of a feature learner $F$, a domain discriminator $D_d$, a feature predictor $P$, two feature translators $T_{s2t}$, $T_{t2s}$ along with their sample discriminators $D_s$ and $D_t$. 

The main goal of our model is to train an effective $F$ which is able to learn domain-invariant feature representations, so that the classifier trained on the source domain can be applied to the target domain. To this end, we introduce the domain discriminator $D_d$ to evaluate the feature quality of $F$ by an adversarial manner. The feature predictor $P$ is further added to condition the domain discriminator $D_d$, so that we can promote the discriminative ability of the learned features and preserve the multimodal structures embedded in the data. However, since the results of $P$ are not always sufficiently accurate, we further argue that truly domain-invariant features should be able to be translated from one domain to the other. Therefore, we train two feature translators $T_{s2t}$, $T_{t2s}$ and learn a cycle-loss. The sample discriminators $D_t$ and $D_s$ are used to evaluate $T_{s2t}$ and $T_{t2s}$, respectively, in an adversarial manner.

For clarity, we show the core ideas of our method in Fig.~\ref{fig:ideaill}. The base network of our 3CATN is the deep feature extractor $F$. Additionally, three adversarial networks are introduced to guide $F$. The first one is the domain discriminator $D_d$ which conditioned by the outputs $P$. The other two are $\{T_{s2t},D_t\}$ and $\{T_{t2s},D_s\}$. Assuming that $T_{t2s}(T_{s2t}(x_s))$ should be similar (or same in an ideal situation) with $x_s$, we calculate a cycle-loss to balance the loss of $D_d$.

\subsection{The Conditional Domain Adversarial Nets}
Since most of the real tasks are multi-class classification rather than binary classification, the features learned from a multi-way classification deep network naturally have multimodal structures~\cite{long2018conditional}. For instance, a liger shares many characteristics of both a lion and a tiger. Thus, the visual features of a liger would have not only its specific structures but also the structures of a lion and a tiger. Such a phenomenon can be reflected by the classifier predictions, e.g., the classifier indicates the probabilities of categorizing a liger image into liger, lion and tiger are 0.8, 0.12 and 0.08, respectively. In other words, the classifier prediction $\mathrm{p}$ carries the possible discriminative information of the multimodal data structures. Inspired by CDAN~\cite{long2018conditional}, we condition both the feature learner $F$ and the domain discriminator $D_d$ with the classifier prediction $\mathrm{p}$. As a result, we have the following minimax game:
\begin{equation}
\label{eq:cdan}
  \begin{array}{l}
 \min\limits_{F,P}\max\limits_{D_d} \mathcal{L}_{con}= ~~-\mathbb{E}[\sum_{c=1}^{C}{\bf 1}_{[y_s=c]}\mathrm{log} \sigma(F(x_s))] \\ 
 ~~~~~~~~~~~~~~~~~~~ \hspace{65pt}+ \lambda (\mathbb{E}[logD_d(\mathcal{\delta}(\mathrm{h}_s))]+\mathbb{E}[log(1-D_d(\mathcal{\delta}(\mathrm{h}_t)))]),
\end{array} 
\end{equation} 
where $\lambda>0$ is a balancing parameter. In this paper, we follow CDAN and fix $\lambda=1$ for fair comparisons. The first term in Eq.~\eqref{eq:cdan} is a supervised cross-entropy loss on the source domain, in which ${\bf 1}_{[\cdot]}$ is an indicator, $\sigma$ is the softmax and $C$ is the possible categories. The second term is a conditional loss which is very similar with conditional GAN~\cite{mirza2014conditional}. It is worth noting that $\mathrm{h=(f,p)}$ is the joint variable of the domain specific features and corresponding classification predictions. $\delta$ is the conditioning strategy defined as follows:
\begin{equation}
\label{eq:delta}
  \begin{array}{l}
 \delta(\mathrm{h})=\begin{cases}
\delta_\otimes \mathrm{(f,p)}& \text{if $\mathrm{dim}_\mathrm{f} \times \mathrm{dim}_\mathrm{p} \le 4096$}\\
\delta_\odot \mathrm{(f,p)}& \text{otherwise,}
\end{cases}
\end{array} 
\end{equation} 
where $\mathrm{dim}$ indicates the dimensionality of a vector. $\delta_\otimes$ is a multilinear map and $\delta_\odot$ is an explicit randomized multilinear map. More details of $\delta_\otimes$ and $\delta_\odot$ can be found in~\cite{long2018conditional}.

\subsection{The Bidirectional Translation}
We can learn a conditional domain adversarial network by optimizing Eq.~\eqref{eq:cdan} which is able to capture the multimodal data structures. However, the success of Eq.~\eqref{eq:cdan} builds on the assumption that the classification prediction $\mathrm{p}$ is sufficiently accurate. The bad news is that such an assumption does not always hold in real-world applications. If the prediction contains significant errors, the errors would be propagated to the feature learner. What is worse, the errors can be magnified with the increase of iterations. 

The ultimate goal of adversarial domain adaptation is to learn domain-invariant features. However, $F$ with flawed conditions is able to fool the domain discriminator but cannot learn truly domain-invariant features. In this paper, we argue that the truly domain-invariant features should be able to be translated from one domain to the other. To this end, we first train a feature translator $T_{s2t}$ to translate the features from the source domain to the target domain. Specifically, we train $T_{s2t}$ in an adversarial fashion. A discriminator $D_t$ is simultaneously trained to distinguish real target domain features from the translated source features. As a result, we have the follow loss function:
\begin{equation}
\label{eq:ts2t1}
  \begin{array}{l}
 \min\limits_{T_{s2t}}\max\limits_{D_t} \mathcal{L}_{s2t}= \mathbb{E}[logD_t(\mathrm{f}_t)]+\mathbb{E}[log(1-D_t(T_{s2t}(\mathrm{f}_s)))].
\end{array} 
\end{equation} 

For convenience, let $\hat{\mathrm f}_t=T_{s2t}(\mathrm{f}_s)$. Since $\hat{\mathrm f}_t$ is translated from $\mathrm{f}_s$, it is expected that $\hat{\mathrm f}_t$ should have the same class information with $\mathrm{f}_s$. Thus, we further introduce a supervised classification loss on $\hat{\mathrm f}_t$ to preserve the semantic consistency and rewritten $\mathcal{L}_{s2t}$ as:
\begin{equation}
\label{eq:ts2t2}
  \begin{array}{l}
 \min\limits_{P,T_{s2t}}\max\limits_{D_t} \mathcal{L}_{s2t}= \mathbb{E}[logD_t(\mathrm{f}_t)]+\mathbb{E}[log(1-D_t(T_{s2t}(\mathrm{f}_s)))] \\ ~~~~~~~~~~~~~~~\hspace{60pt}- \beta \mathbb{E}[\sum_{c=1}^{C}{\bf 1}_{[\hat{\mathrm y}_t=c]}\mathrm{log} \sigma(\hat{\mathrm f}_t)].
\end{array} 
\end{equation}

Similarly, we can also train a mapping from the the other direction. Specifically, we further train $T_{t2s}$ in an adversarial fashion. A discriminator $D_s$ is simultaneously trained to distinguish real source domain features from the translated target features. As a result, we have the follow loss function:
\begin{equation}
\label{eq:tt2s}
  \begin{array}{l}
 \min\limits_{T_{t2s}}\max\limits_{D_s} \mathcal{L}_{t2s}= \mathbb{E}[logD_s(\mathrm{f}_s)]+\mathbb{E}[log(1-D_s(T_{t2s}(\mathrm{f}_t)))].
\end{array} 
\end{equation} 

\subsection{The Cycle-consistent Loss}
Inspired by the recent work on image-to-image translation~\cite{zhu2017unpaired}, we encourage our model to preserve the original data information. For instance, preserving the information of $\mathrm{f}_s$ when translating it to $\hat{\mathrm f}_t$. Let us explain this consideration by an example. The handwritten digit~$1$ can be similar to~$7$ if the upper part of~$1$ is distorted during the translation. However, if we preserve the original information of~$1$, such a error would not happen. In our model, therefore, we deploy a cycle-consistent loss to keep the consistency of original features before and after the translation. Formally, we expect that $T_{t2s}(T_{s2t}(\mathrm{f}_s))\approx \mathrm{f}_s$ and $T_{s2t}(T_{t2s}(\mathrm{f}_t))\approx \mathrm{f}_t$. As a result, we define the cycle-loss as:
\begin{equation}
\label{eq:tt2s}
  \begin{array}{l}
 \min\limits_{T_{t2s},T_{t2s}} \mathcal{L}_{cyc}= \mathbb{E}[\|T_{t2s}(T_{s2t}(\mathrm{f}_s))-\mathrm{f}_s\|_2^2] \\ \hspace{60pt}+\mathbb{E}[\|T_{s2t}(T_{t2s}(\mathrm{f}_t))-\mathrm{f}_t\|_2^2],
\end{array} 
\end{equation} 

\subsection{Overall Objective Function}
By considering above all discussions on conditional adversarial training, bidirectional translation and cycle-consistency, we have the following loss for our 3CATN model:
\begin{equation}
\label{eq:loss}
  \begin{array}{l}
 \mathcal{L}_{3CATN}= \mathcal{L}_{con}+\eta_1(\mathcal{L}_{s2t}+\mathcal{L}_{t2s})+\eta_2\mathcal{L}_{cyc},
\end{array} 
\end{equation} 
where $\eta_1, \eta_2>0$ are two balancing parameters to control the contribution of each part. Since we claim that truly domain-invariant features should be able to be translated from one domain to the other, we deploy the same weight for $\mathcal{L}_{z2t}$ and $\mathcal{L}_{t2s}$ in this paper to show the equality of the bidirectional translations. With the overall loss, we have our overall objective function:
\begin{equation}
\label{eq:objective}
  \begin{array}{l}
 \min\limits_{F,P,T_{t2s},T_{t2s}}~~~\max\limits_{D_d,D_s,D_t} \mathcal{L}_{3CATN}.
\end{array} 
\end{equation} 


Once the model is well-trained, we can deploy the feature learner $F$ to learn domain invariant features for both the source and target domains. Then, the classifier trained on $F(X_s)$ can be directly used to handle $F(X_t)$.

\section{Experiments}

\subsection{Datasets Description}
{\bf MNIST}, {\bf USPS} and Street View House Numbers ({\bf SVHN})~\cite{netzer2011reading}, are three widely used handwritten digits dataset. MNIST consists of 60,000 training samples and 10,000 test samples. USPS is comprised of 7,291 training samples and 2,007 test samples. SVHN contains over 600,000 labeled digits cropped from Street View images. 

 {\bf Office-31}~\cite{saenko2010adapting} is the most popular benchmark in domain adaptation. It consists of three subsets, i.e., {\bf Amazon} (A), {\bf Webcam} (W) and {\bf DSLR} (D). Specifically, the images in amazon is downloaded from amazon.com. Webcam and DSLR contain images captured by a web camera and a digital SLR camera, respectively. In total, there are 4,652 samples from 31 categories in office-31 dataset. 


 {\bf VisDA}~\cite{visda2017} classification dataset is the currently largest dataset in the community. It consists of a training domain, a validation domain and a test domain. In total, the VisDA dataset contains over 280K images from 12 classes. In this paper, we follow the same settings in previous works~\cite{saito2018maximum,pinheiro2018unsupervised} and use the training set as the source domain and the validation domain as the target domain. 

 \subsection{Experiment Protocols}
Our model mainly consists of two parts. The first part is inspired by CDAN~\cite{long2018conditional} to take advantage of conditioning with classifier predictions. However, it is easy to figure out that an inaccurate condition will lead to negative results. Thus, the second part of our model is a cycle-consistent translation to prevent the negative results. We argue that truly domain-invariant features should be able to be translated from one domain to the other. Apparently, CDAN can be regarded as a baseline of our method. Furthermore, CDAN is published very recently. The results of CDAN represent state-of-art performance in the community. As a result, we mainly compare our method with CDAN. For fair comparisons, we keep exactly the same experiment protocols with CDAN~\cite{long2018conditional}. Our codes are available at github\footnote{github.com/lijin118/3CATN}.

In this paper, we deploy the standard unsupervised domain adaptation settings which were widely used in previous work~\cite{long2018conditional,gong2012geodesic}. Labeled source domain samples $\{X_s,Y_s\}$ and unlabeled target domain samples $X_t$ are used for training. The reported results are the classification accuracy on target samples:
$\displaystyle accuracy = {{|x:~x \in X_t ~\wedge~ \hat{y_t}=y_t|}/{|x:~ x \in X_t|}}$, where ${\hat{y_t}}$ is the predicted label of the target domain generated by our model, and ${y}_t$ is the ground truth label vector. 

For digits recognition on MNIST, USPS and SVHN, we use the same setting in CDAN. Specifically, $60000$, $7291$ and $73257$ images from MNIST, USPS and SVHN, respectively, are used for training. The basic network structure for digits recognition is similar with LeNet. We set the batch size to $224$ and the learning rate as $10^{-3}$.

For object recognition on Office-31, we also deploy the same setting in CDAN. The basic network for feature extraction is ResNet-50~\cite{he2016deep} which is pre-trained on ImageNet. The batch size is 32 and the learning rate is $10^{-3}$. We optimize our model by mini-batch stochastic gradient descent with a weight decay of $5\times10^{-4}$ and momentum of $0.9$. For the experiments on VisDA 2017, we also deploy ResNet-50 as the base architecture.

The feature translators are implemented by one fc layer and three conv layers. The discriminator $D_s$ and $D_t$ are implemented by two conv layers and one fc layer. The domain discriminator $D_d$ is implemented by three fc layers. All the hyper-parameters of this paper are tuned by importance weighted cross validation~\cite{sugiyama2007covariate}. The sensitivity of parameters are discussed later in this section. 

Traditional methods with deep features: TCA~\cite{pan2011domain} and GFK~\cite{gong2012geodesic}. Deep CNN methods: DANN~\cite{ganin2014unsupervised}, DDC~\cite{tzeng2014deep}, DAN~\cite{long2015learning}, DRCN~\cite{ghifary2016deep}, JAN~\cite{long2017deep} and SimNet~\cite{pinheiro2018unsupervised}. GAN methods: ADDA~\cite{tzeng2017adversarial}, JAN-A~\cite{long2017deep}, RevGrad~\cite{ganin2016domain}, CoGAN~\cite{liu2016coupled}, MCD~\cite{saito2018maximum} and CDAN~\cite{long2018conditional} are used for comparison. The results of some compared methods are cited from JAN and CDAN.

\subsection{Digits Recognition}

\begin{table}[t!p]
\centering
\caption{Results (\%) of digits recognition. The best results are highlighted by bold numbers. The compared methods have the same experimental settings except for UNIT on SVHN $\rightarrow$MNIST which uses a larger training set. The results of the baselines are cited from corresponding papers.}
\vspace{-8pt}
\label{tab:digits}
\setlength{\tabcolsep}{0.3mm}{
\begin{tabular}{lccc}
\toprule
Method & MNIST$\rightarrow$USPS & USPS$\rightarrow$MNIST & SVHN $\rightarrow$MNIST  \\
\midrule
Source only & $82.2 \pm 0.8$ & $69.6 \pm 3.8$ & $67.1 \pm 0.6$    \\
DANN~\cite{ganin2014unsupervised} & - & - & $73.6$    \\
ADDA~\cite{tzeng2017adversarial}  & $89.4 \pm 0.2$ & $90.1 \pm 0.8$ & $76.0 \pm 1.8$   \\
DRCN~\cite{ghifary2016deep} & $91.8 \pm 0.1$ & $73.7 \pm 0.1$ & $82.0 \pm 0.1$   \\     
RevGrad~\cite{ganin2016domain} & $89.1 \pm 0.2$ & $89.9 \pm 0.3$ & $-$   \\
CoGAN~\cite{liu2016coupled} & $91.2$ & $89.1$ & $-$  \\
UNIT~\cite{liu2017unsupervised} & $95.9$ & $93.6$ & $90.5$  \\
CDAN~\cite{long2018conditional} & $95.6$ & $98.0$ & $89.2$  \\ 
\midrule
{3CATN [Ours]}  & ${\bf 96.1 \pm 0.2}$ & ${\bf 98.3 \pm 0.2}$ & ${\bf 92.5 \pm 0.3}$  \\
\midrule
{Target Supervised} & 96.3 & 99.2 & 99.2 \\
\bottomrule
\end{tabular}}
\end{table}

The results of hand-written digits recognition are reported in Table~\ref{tab:digits}. We also report the results on source only as a baseline. The source only means the model is trained on only the source data and then directly applied on the target data. The other compared methods are mainly based on GANs. From the results we can see that our method is able to outperform previous state-of-the-arts. Specifically, we achieved 0.5\%, 0.3\% and 3.3\% accuracy improvement on MNIST$\rightarrow$USPS, USPS$\rightarrow$MNIST and SVHN $\rightarrow$MNIST, respectively. It is worth noting that the improvement is very hard to achieve since previous state-of-the-arts are very close to the results of target full supervised setting. Nevertheless, we achieve a significant improvement on the hardest task SVHN $\rightarrow$MNIST.

CDAN performs very well when the classifier predictions are sufficiently accurate. The results on the hand-written digits are already sufficiently accurate for most recent works. Thus, our advantages are not quite clear on these evaluations. We perform comparably with the target supervised model. As we claimed before, our model would significantly improve CDAN when the classifier perditions are inaccurate. This is somewhat proved by the results on SVHN $\rightarrow$MNIST. It will be further verified in following evaluations. In conclusion, our method is able to take full advantage of classifier predictions condition when they are accurate. At the same time, we can also avoid the negative effects when the classifier conditions are inaccurate. Our approach achieves a dynamic balance by maximizing the advantages and minimizing the shortages.

It is also worth noting that the entropy conditioning in CDAN is designed to handle the issue of inaccurate classifier predictions. However, sample entropy does not necessarily indicate the classification accuracy. A strong evidence is that CDAN+E (CDAN with entropy conditioning) does not show significant advantage against vanilla CDAN. For instance, the average accuracies of CDAN+E and CDAN on Office-31 are 87.7\% and 86.6\%, respectively. The results of CDAN reported in this paper are all from CDAN+E since it is generally better than CDAN. We did not specific CDAN+E from CDAN for the sake of simplicity and clarity.

\begin{table*}[ht!p]
\centering
\caption{Domain adaptation results (accuracy~\%) on Office-31. Where average~1 is the overall average and average~2 is the average over 4 challenging evaluations except for W$\rightarrow$D and D$\rightarrow$W. The best results are highlighted by bold numbers.}
\vspace{-8pt}
\label{tab:office}
\setlength{\tabcolsep}{2.8mm}{
\begin{tabular}{lcccccc|cc}
\toprule
Method & A$\rightarrow$D & A$\rightarrow$W & D$\rightarrow$A & D$\rightarrow$W & W$\rightarrow$A & W$\rightarrow$D & Average~1 & Average~2\\
\midrule
ResNet~\cite{he2016deep} & $68.9 \pm 0.2$~ & $ 68.4 \pm 0.2$~ & $62.5 \pm 0.3$~ & $96.7 \pm 0.1$~ & $60.7 \pm 0.3$~ & $99.3 \pm 0.1$~ & $76.1$ & $65.1$\\ 
TCA~\cite{pan2011domain}   & $74.1 \pm 0.0$ & $72.7 \pm 0.0$ & $61.7 \pm 0.0$ & $96.7 \pm 0.0$ & $60.9 \pm 0.0$ & $99.6 \pm 0.0$ & $77.6 $ & $67.4$ \\
GFK~\cite{gong2012geodesic} & $74.5 \pm 0.0$ & $72.8 \pm 0.0$ & $63.4 \pm 0.0$ & $95.0 \pm 0.0$ & $61.0 \pm 0.0$ & $98.2 \pm 0.0$  & $77.5$ & $67.9$\\  
DDC~\cite{tzeng2014deep} & $76.5 \pm 0.3$ & $75.6 \pm 0.2$ & $62.2 \pm 0.4$ & $96.0 \pm 0.2$ & $61.5 \pm 0.5$ & $98.2 \pm 0.1$ & $78.3$ & $69.0$ \\
DAN~\cite{long2015learning} & $78.6 \pm 0.2$ & $80.5 \pm 0.4$ & $63.6 \pm 0.3$ & $97.1 \pm 0.2$ & $62.8 \pm 0.2$ & $99.6 \pm 0.1$ & $80.4$ & $71.4$ \\
RevGrad~\cite{ganin2016domain} & $79.7 \pm 0.4$ & $ 82.0 \pm 0.4$ & $68.2 \pm 0.4$ & $96.9 \pm 0.2$ & $67.4 \pm 0.5$ & $99.1 \pm 0.1$ & $82.2$ & $74.3$\\ 
MCD~\cite{saito2018maximum} & $74.5 \pm 0.6$ & $68.3 \pm 0.2$& $49.9 \pm 0.5$ & $90.7 \pm 0.8$ & $43.5 \pm 0.5$ & $98.3 \pm 0.5$ & $70.9 $ & $59.1$\\ 
JAN~\cite{long2017deep} & $84.7 \pm 0.3$ & $85.4 \pm 0.3$& $68.6 \pm 0.3$ & ${ 97.4 \pm 0.2}$ & $70.0 \pm 0.4$ & ${ 99.8 \pm 0.2}$ & $84.3$ & $77.2$\\ 
JAN-A~\cite{long2017deep} & ${85.1 \pm 0.4}$ & ${ 86.0 \pm 0.4}$& ${ 69.2 \pm 0.4}$ & $96.7 \pm 0.3$ & ${ 70.7 \pm 0.5}$ & $99.7 \pm 0.1$ & ${ 84.6}$ & ${ 77.8}$\\  
GTA~\cite{sankaranarayanan2018generate} & ${ 87.7 \pm 0.5}$ & ${ 89.5 \pm 0.5}$& ${ 72.8 \pm 0.3}$ & $97.9 \pm 0.3$ & ${ 71.4 \pm 0.4}$ & $99.8 \pm 0.4$ & ${86.5}$ & ${80.3}$\\
CDAN~\cite{long2018conditional} & ${ 92.9 \pm 0.2}$ & ${ 94.1 \pm 0.1}$& ${ 71.0 \pm 0.3}$ & $98.6 \pm 0.1$ & ${ 69.3 \pm 0.3}$ & $100 \pm 0.0$ & ${87.7}$ & ${81.8}$\\  
\midrule
3CATN [Ours]  & ${\bf 94.1 \pm 0.3}$ & ${\bf 95.3 \pm 0.2}$ & ${\bf 73.1 \pm 0.2}$ & ${\bf 99.3 \pm 0.5}$ & ${\bf 71.5 \pm 0.6}$ & ${\bf 100 \pm 0.0}$ & ${\bf 88.9}$ & {\bf 83.5}\\
\bottomrule
\end{tabular}}
\end{table*}


\subsection{Objects Recognition on Office-31}
The results of different methods on Office-31 are reported in Table~\ref{tab:office}. It is worth noting that we report two average results, e.g., Average~1 and Average~2. Average~1 is the average over all six evaluations. Average~2 is the average over four evaluations which excludes W$\rightarrow$D and D$\rightarrow$W since the two have very high accuracy for almost every method. As a result, Average~2 can reflect the performance on relatively harder tasks. From the results, we can have the following observations:

1) Transfer learning methods perform much better than the baseline ResNet-50 which is trained on only the source domain data. It proves that domain adaptation is able to minimize the domain shift and it is practical for real-world applications.

2) End-to-end deep methods outperform two-step methods which split feature extraction and knowledge adaptation. By simultaneously optimizing the two parts, the feature learner is able to receive the feedback from latter layers. The feedback can benefit the feature learner to learn more domain-invariant features.

3) Compared with previous end-to-end deep models, adversarial domain adaptation achieves state-of-the-art performance recently. Comparing JAN with CDAN, although both of them use the same base architecture, it is obvious that a conditional domain discriminator can significantly improve the domain adaptation performance.

4) Comparing average~1 and average~2, we can see that most methods have a nearly overall average accuracy since the results on W$\rightarrow$D and D$\rightarrow$W are very high, the two soft the overall average on tough tasks. However, the baseline ResNet also performs very well on the two evaluations. Thus, average~2 is better to reflect the performance on hard tasks of Office-31. Regarding average~2, we can see that CDAN really improves state-of-the art, which verifies that leveraging the condition with classifier prediction is quite effective. Our method is also benefited from the classifier predictions. At the same time, we can also observe that CDAN performs remarkably on A$\rightarrow$D and A$\rightarrow$W where the classifier predictions are relatively high (over 80\% for JAN). However, for the cases D$\rightarrow$A and W$\rightarrow$A where the classifier predictions are relatively poor (below 70\% for JAN), CDAN has no advantage against previous state-of-the-art approaches. It is even worse than GTA on both D$\rightarrow$A and W$\rightarrow$A. This comparison exactly verifies that the condition with classifier predictions is not reliable when the predictions are inaccurate.

5) Although we deployed the same condition with CDAN, our model is able to pervert the negative effects caused by inaccurate classifier conditions. We argue that truly domain-invariant features should be able to translate from one domain the other. As a result, we introduce two feature translation losses and one cycle-consistent loss. The results verify that our model can perform well with good classifier conditions. At the same time, our model is also robust with bad classifier conditions. This verifies our claim that our approach is able to not only capture the complex multimodal structures but also avoid the negative effects caused by uncertain conditions.

6) In terms of numbers, we achieved 1.2\% accuracy improvement w.r.t. average~1 and 1.7\% accuracy improvement w.r.t. average~2. It is worth noting that we outperform state-of-the-art CDAN 2.1\% and 2.2\% on the two hardest tasks D$\rightarrow$A and W$\rightarrow$A, respectively.

\subsection{Large-scale Test}
We further evaluate our model on large-scale domain adaptation datasets VisDA 2017. It is worth noting that VisDA is more challenging than Office-31. Compared with Office-31, VisDA has much more images. It is also worth noting that VisDA challenge has classification and segmentation tasks. Limited by space, in this paper we only focus on the classification task. The classification results on VisDA are reported in Table~\ref{tab:visda1}.


The results on VisDA also verify that our method can outperform previous state-of-the-art with significant advantages. We achieve 3.2\% accuracy improvement over the $12$ classes in VisDA. From the results on VisDA, we can see that our method is able to perform well on large-scale datasets. It is able to maximize the advantages of accurate predictions and, at the same time, prevent the negative effects caused by inaccurate predictions.

\begin{table*}[ht!p]
\centering
\caption{Domain adaptation results (accuracy~\%) on VisDA-2017 dataset. All of the base networks are ResNet-50. }
\vspace{-8pt}
\label{tab:visda1}
\begin{tabular}{ccccccccc}
\toprule
Method & ~ResNet\cite{he2016deep}~ & ~DAN~\cite{long2015learning}~ & ~RTN~\cite{long2016unsupervised}~ & ~RevGrad~\cite{ganin2016domain}~  & ~JAN~\cite{long2017deep}~ & ~SimNet~\cite{pinheiro2018unsupervised}~ & CDAN~\cite{long2018conditional} & ~3CATN [Ours]~\\
\midrule
Result & 49.5 & 53.0 & 53.6 & 55.0  & 61.6 & 69.6 & 70.0 & {\bf 73.2} \\ 
\bottomrule
\end{tabular}
\end{table*}

\begin{figure*}[t!h]
\begin{center}
\subfigure[Classification loss ($\beta$)]{
\includegraphics[width=0.23\linewidth]{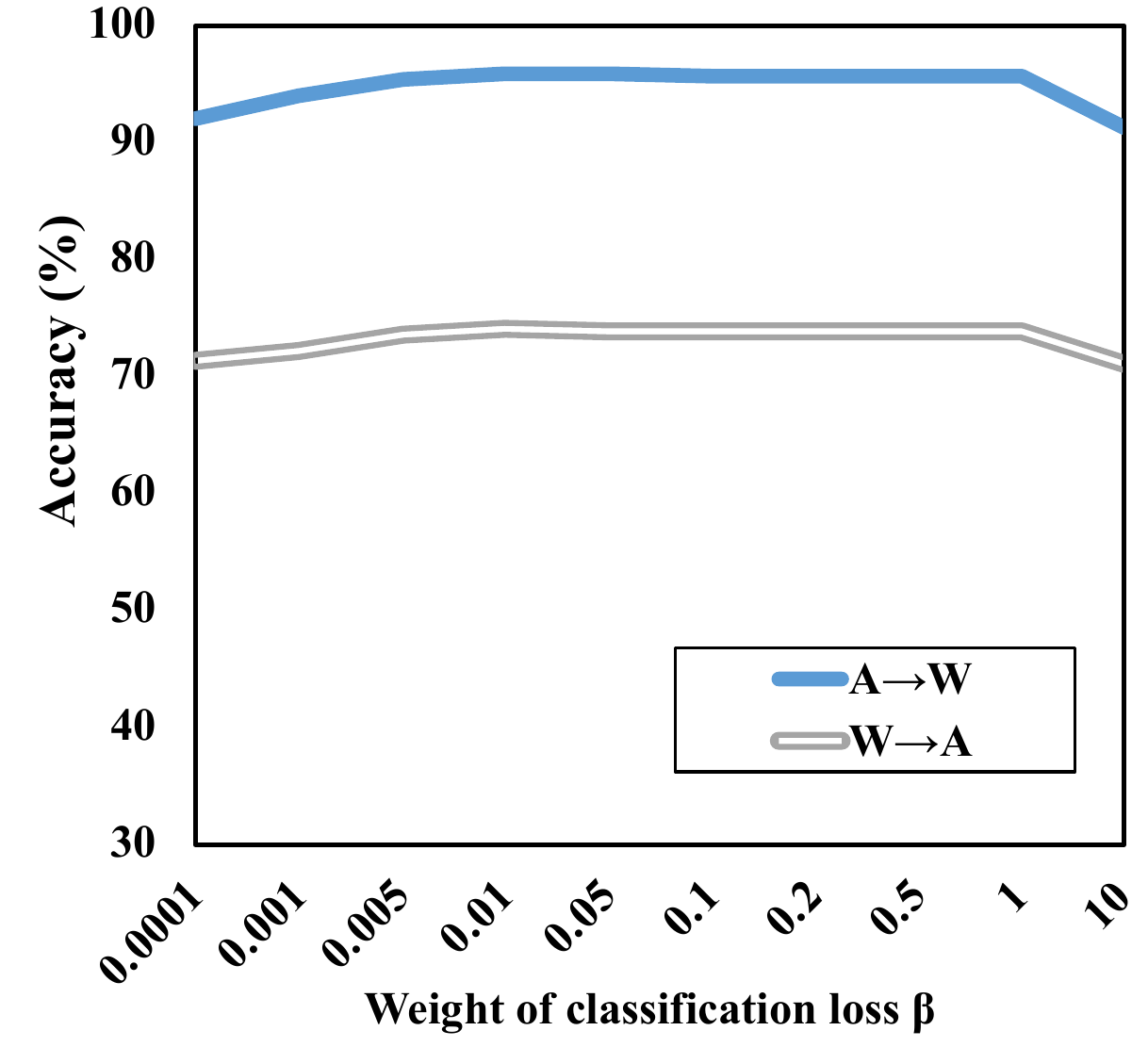}
}
\subfigure[Cycle-loss ($\eta_1$)]{
\includegraphics[width=0.232\linewidth]{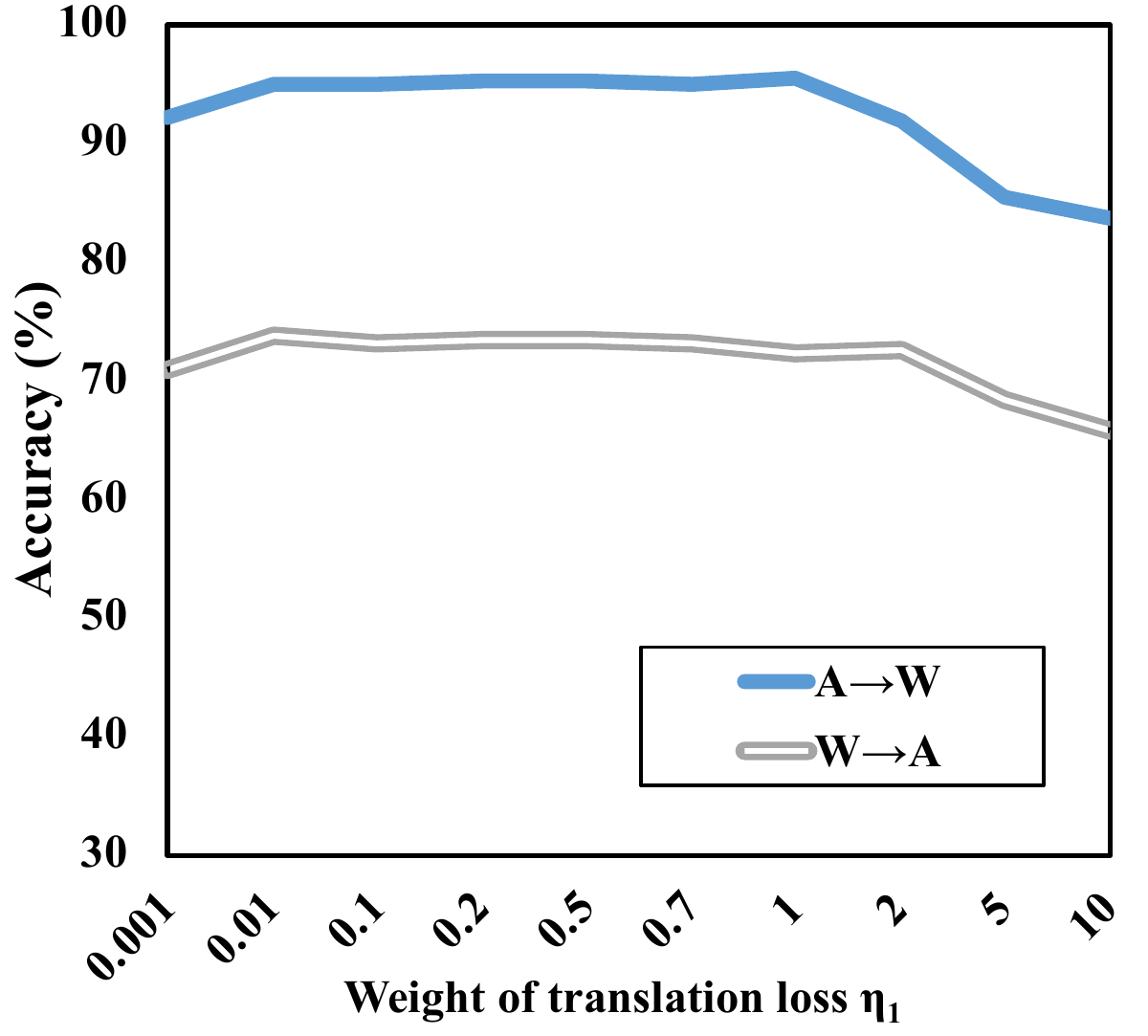}
}
\subfigure[Translation loss ($\eta_2$)]{
\includegraphics[width=0.23\linewidth]{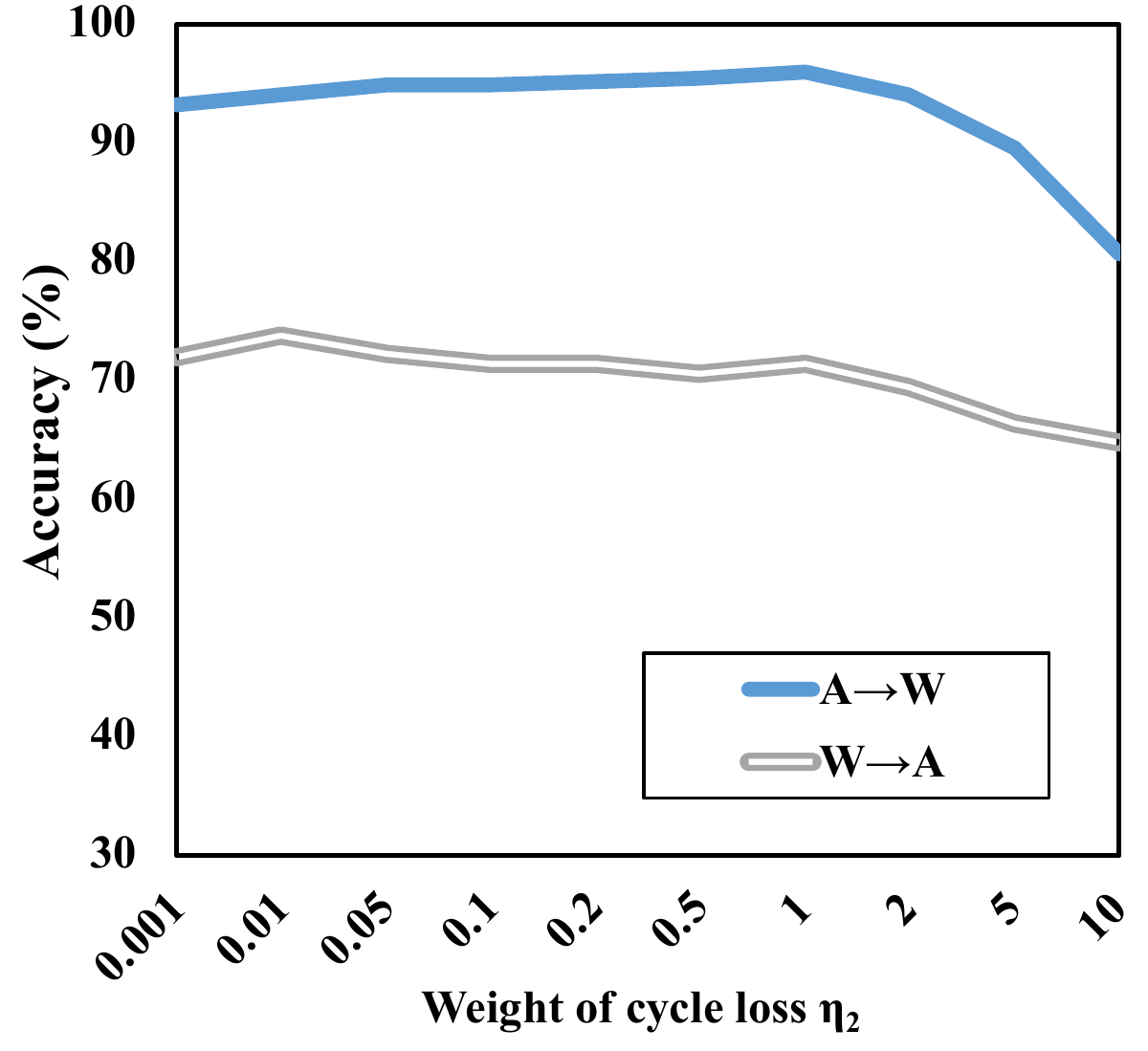}
}
\subfigure[Training stability]{
\includegraphics[width=0.238\linewidth]{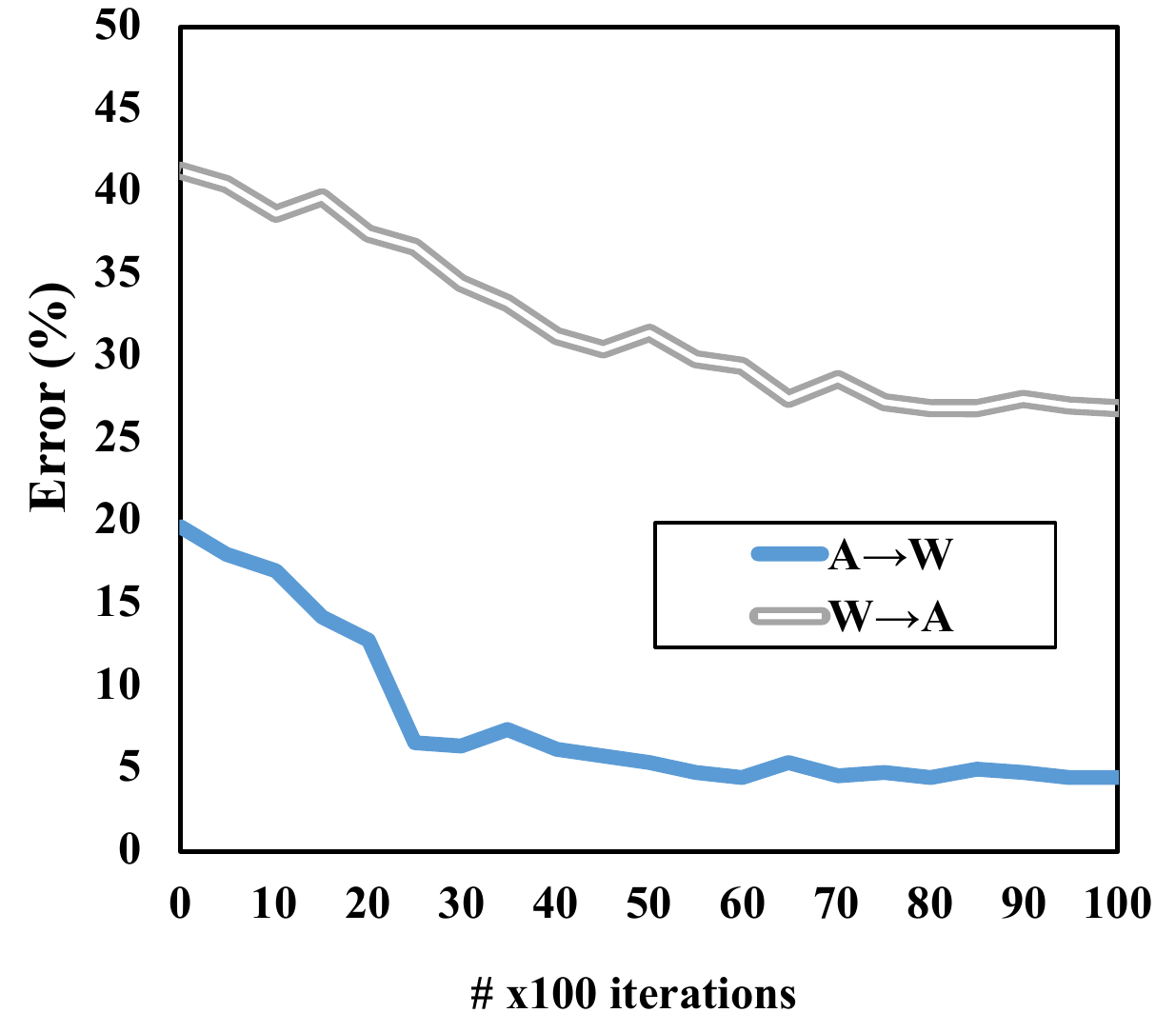}
}
\end{center}
\vspace{-16pt}
\caption{Parameter sensitivity (a-c) and training stability (d). The parameters in our model are tuned by importance weighted cross validation~\cite{sugiyama2007covariate}. W$\rightarrow$A and A$\rightarrow$W on Office-31 dataset are evaluated as examples. }
\label{fig:para}
\vspace{-5pt}
\end{figure*}

\begin{figure*}[t!h]
\begin{center}
\includegraphics[width=0.95\linewidth]{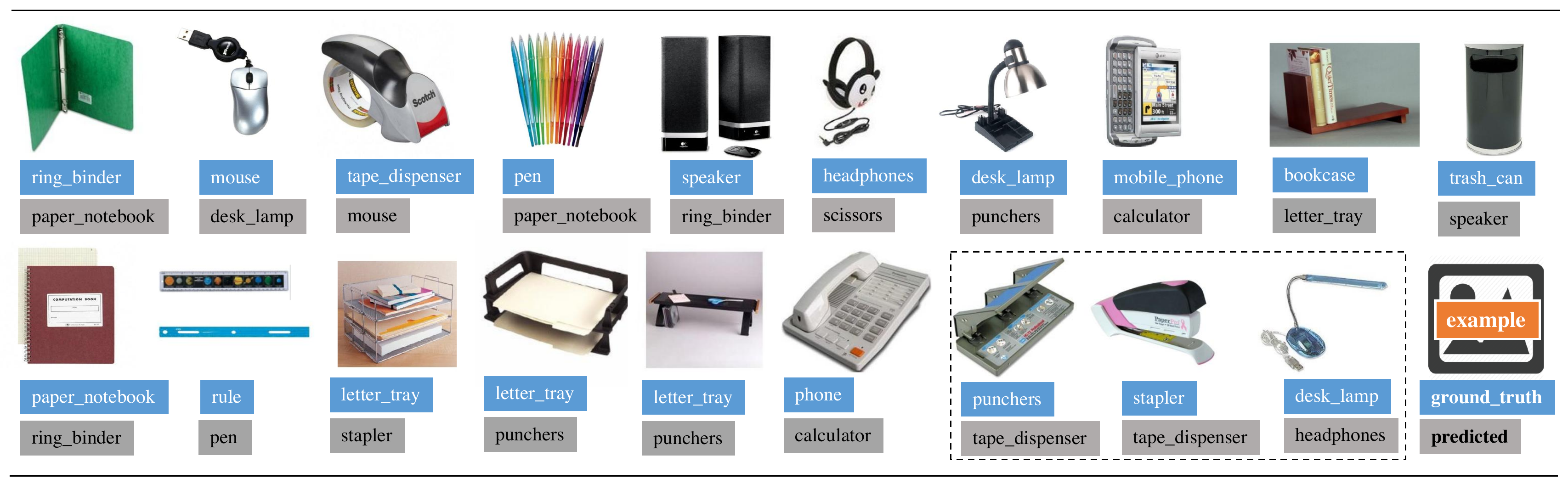}
\end{center}
\vspace{-10pt}
\caption{Qualitative results on W$\rightarrow$A. Blue label is the ground-truth and gray label is the prediction. This figure mainly shows some randomly selected samples which are wrongly predicted by CDAN but correctly classified by our 3CATN. Although almost all of the samples which are correctly predicted by CDAN are also correctly handled by our method, we show several exceptions within the dash lines for the sake of fairness.}
\label{fig:samples}
\vspace{-5pt}
\end{figure*}

\subsection{Qualitative Evaluation}
For a better understanding, we conduct a qualitative evaluation on W$\rightarrow$A. We show some images which are misclassified by CDAN but well-handled by our method in Fig.~\ref{fig:samples}. On the evaluation of W$\rightarrow$A, there are 163 out of 2849 samples are misclassified by CDAN but well-handled by our 3CATN. At the same time, almost all the correctly classified samples by CDAN are also correctly classified by our 3CATN. There are only 15 exceptions. For the sake of fairness, we also show some of them in Fig.~\ref{fig:samples}.

\subsection{Model Analysis}
{\bf Parameters.} The parameters in our model are tuned by importance weighted cross validation~\cite{sugiyama2007covariate}. In most evaluations, we fix $\lambda=1$ as in CDAN. Since $\beta$ plays the similar role with $\lambda$, we also set $\beta=1$. The weight for feature translators $\eta_1$ is set as $0.01$ and the weight for cycle-loss $\eta_2$ is set as $0.1$. To fully evaluate our model, we report the parameter sensitivities of $\beta$, $\eta_1$ and $\eta_2$ in Fig.~\ref{fig:para}(a), Fig.~\ref{fig:para}(b) and Fig.~\ref{fig:para}(c), respectively. It can be seen that $\beta$ is not sensitive. Our model performs stably with different values of $\beta$. $\eta_1$ and $\eta_2$ are suggested to be set smaller than~1.

\noindent {\bf Stability.} Adversarial training is famous for hard of training. Therefore, we report the training stability of our model in Fig.~\ref{fig:para}(d). It can be seen that 3CATN performs stably with the increase of iterations and it usually can achieve a stable result within $10,000$ iterations.

\noindent {\bf Ablation.} To show the effect of each component in our model, we report the results of ablation study in Table~\ref{tab:ablation}. The basic network of our model is the ResNet which learns new feature representations. The result of ResNet is reported as S0. Then, we leverage the conditional adversarial training to capture the multimodal data embedded in the data. The result is reported as S1. Furthermore, two feature translators are introduced to learn truly domain-invariant features and balance the inaccurate conditioning issue, which is reported as S2. At last, a cycle-consistent loss is calculated to preserve the consistency before and after feature translation. The result is reported as S3. The results in Table~\ref{tab:ablation} verify that each part in our model plays an important role for the final results. 

It is worth noting that S1 is based on CDAN, S2 and S3 are our contributions. Combining the results in Table~\ref{tab:ablation} and Fig.~\ref{fig:samples}, we can find that CDAN is risky when the classifier predictions are not sufficiently accurate, e.g., the classifier may be confused with a phone and a calculator and putting a condition of calculator on the category phone leads to the misclassification of phone images. Our method, however, is able to alleviate the negative effects caused by inaccurate conditions. In the introduction, we argued that truly domain-invariant features should be able to be translated from one domain to the other. To this end, we introduce two feature translators and one cycle-consistent loss. The results of ablation study and qualitative evaluation verify that both translators and cycle-loss definitely benefit the model. For instance, we can see that feature translators and cycle-loss boost the overall performance on D$\rightarrow$A with 0.9\% and 1.2\%, respectively.

\begin{table}[t!p]
\centering
\caption{The results of ablation study on D$\rightarrow$A and W$\rightarrow$A. }
\vspace{-5pt}
\label{tab:ablation}
\small \begin{tabular}{lcc}
\toprule
Settings & D$\rightarrow$A & W$\rightarrow$A  \\
\midrule
S0: ResNet  & 62.5 & 60.7   \\
S1: S0+ conditional transfer loss    & 71.0 & 69.3   \\
S2: S1 + feature translation loss & 71.9 & 70.4   \\
S3: S2 + cycle-consistent loss  & {\bf 73.1} & {\bf 71.5}  \\     
S4: S3 - conditional transfer loss   & 67.5 & 66.9  \\     
\bottomrule
\end{tabular}
\vspace{-10pt}
\end{table}

\section{Conclusion}
In this paper, we proposed a novel adversarial domain adaptation method 3CATN. 3CATN takes advantage of conditional adversarial training. In order to address the inaccurate conditioning issue, we argue that truly domain-invariant features should be able to be translated from one domain to the other. As a result, two feature translators and one corresponding cycle-consistent loss are introduced into conditional adversarial domain adaptation networks. Extensive experiments verify that our method is able to outperform previous state-of-the-art methods with significant advantages. In our model, we address the inaccurate conditioning issue with external forces, e.g., balancing the weight between classifier condition and feature translation. In our further work, we will study handling the problem from internal side, i.e., explicitly evaluating the accuracy of classifier predictions.

\section{Acknowledgments}
This work was supported in part by the NSFC under Grant 61806039 and 61832001, in part by ARC under DP190102353, in part by the National Postdoctoral Program for Innovative Talents under Grant BX201700045, in part by the China Postdoctoral Science Foundation under Grant 2017M623006 and in part by Sichuan Department of Science and Technology under Grant 2019YFG0141.

%
\balance
\bibliographystyle{ACM-Reference-Format}
\bibliography{mm19}


\end{document}